\documentclass[letterpaper, 10 pt, journal, twoside]{IEEEtran}
\usepackage{amsmath,amsfonts}
\usepackage{algorithmic}
\usepackage{algorithm}
\usepackage{array}
\usepackage[caption=false,font=normalsize,labelfont=sf,textfont=sf]{subfig}
\usepackage{textcomp}
\usepackage{stfloats}
\usepackage{url}
\usepackage{verbatim}
\usepackage{graphicx}
\usepackage{cite}
\usepackage{cite}
\usepackage{amsmath,amssymb,amsfonts}
\usepackage{algorithmic}
\usepackage{graphicx}
\usepackage{textcomp}
\usepackage{xcolor}
\usepackage{inlinelabel}
\usepackage{tabularx}
\usepackage{float}
\usepackage{multirow}
\usepackage{multicol}
\usepackage{booktabs} 
\usepackage{diagbox}
\usepackage{array}
\usepackage{color}
\usepackage{hhline}
\usepackage{caption}
\usepackage{titlesec}
\usepackage{url}
\usepackage{hyperref}

\begin{document}

\title{
\LARGE \bf
Adaptive Wiping: Adaptive contact-rich manipulation \\ through few-shot imitation learning with Force-Torque feedback \\ and pre-trained object representations
}


\author{Chikaha Tsuji$^{1*}$, Enrique Coronado$^{2}$, Pablo Osorio$^{3}$ and Gentiane Venture$^{4,2*}$
\thanks{$^{1}$ Department of Mechano-Informatics, The University of Tokyo, Japan. tsujchik@g.ecc.u-tokyo.ac.jp}%
\thanks{$^{2}$ National Institute of Advanced Industrial Science and Technology, Japan}%
\thanks{$^{3}$ Department of Mechanical Systems Engineering, Tokyo University of Agriculture and Technology, Japan. }%
\thanks{$^{4}$ Department of Mechanical Engineering, The University of Tokyo, Japan. venture@g.ecc.u-tokyo.ac.jp *Corresponding authors}%
}


\maketitle

\begin{abstract}
Imitation learning offers a pathway for robots to perform repetitive tasks, allowing humans to focus on more engaging and meaningful activities. However, challenges arise from the need for extensive demonstrations and the disparity between training and real-world environments. This paper focuses on contact-rich tasks like wiping with soft and deformable objects, requiring adaptive force control to handle variations in wiping surface height and the sponge's physical properties.
To address these challenges, we propose a novel method that integrates real-time force-torque (FT) feedback with pre-trained object representations. This approach allows robots to dynamically adjust to previously unseen changes in surface heights and sponges' physical properties.
In real-world experiments, our method achieved 96\% accuracy in applying reference forces, significantly outperforming the previous method that lacked an FT feedback loop, which only achieved 4\% accuracy. To evaluate the adaptability of our approach, we conducted experiments under different conditions from the training setup, involving 40 scenarios using 10 sponges with varying physical properties and 4 types of wiping surface heights, demonstrating significant improvements in the robot's adaptability by analyzing force trajectories. \href{https://sites.google.com/view/adaptive-wiping}{https://sites.google.com/view/adaptive-wiping}
\end{abstract}


\begin{figure}[t]
\centerline{\includegraphics[scale=0.72]{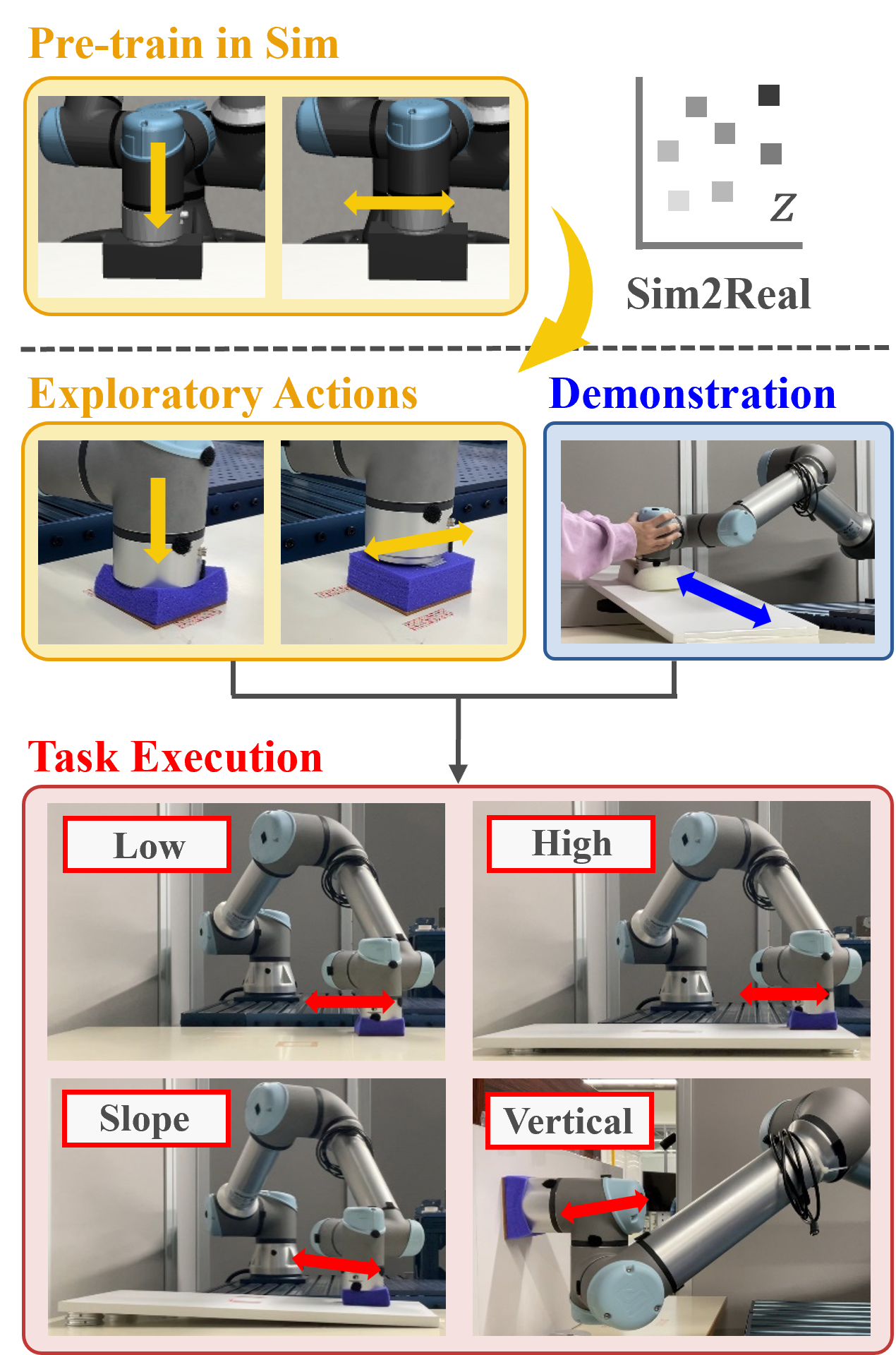}}
\vspace*{1.5mm}
\caption{Wiping Experiments: Pre-trained sponge properties in simulation (top), collected real-world data via exploratory actions and human demonstrations (middle), and tested 40 scenarios with 10 sponges, including 9 unseen sponges, and 4 surface heights, including a wall (bottom).}
\label{fig:task}
\vspace*{-2mm}
\end{figure}


\section{Introduction}
\IEEEPARstart{R}{obots} are crucial for handling mundane tasks, but pre-programming each task is impractical, leading to increased interest in imitation learning~\cite{hussein2017imitation}. Despite its benefits, challenges like the need for extensive demonstrations and discrepancies between training and real-world environments persist~\cite{NIPS2017_ba386660}. Thus, robots must not merely mimic but adapt to new environments, even with limited demonstration data.

A challenging aspect of robotic manipulation is executing contact-rich tasks, which involve extensive physical interactions. Interestingly, those involving deformable objects pose particular challenges due to the need for precise force
control and adaptation to changes \cite{doi:10.1126/science.aat8414}. Wiping tasks, for example, demand careful force adjustments based on wiping surface height and sponge's physical properties.

Therefore, in this paper, we address the challenge – \emph{Could robots learn a versatile manipulation policy via few-shot imitation learning capable of adapting to environmental changes: the height of manipulating surface and the physical properties of manipulated objects?}

\begin{figure*}[t]  
 \begin{center}
  \includegraphics[scale=0.37]{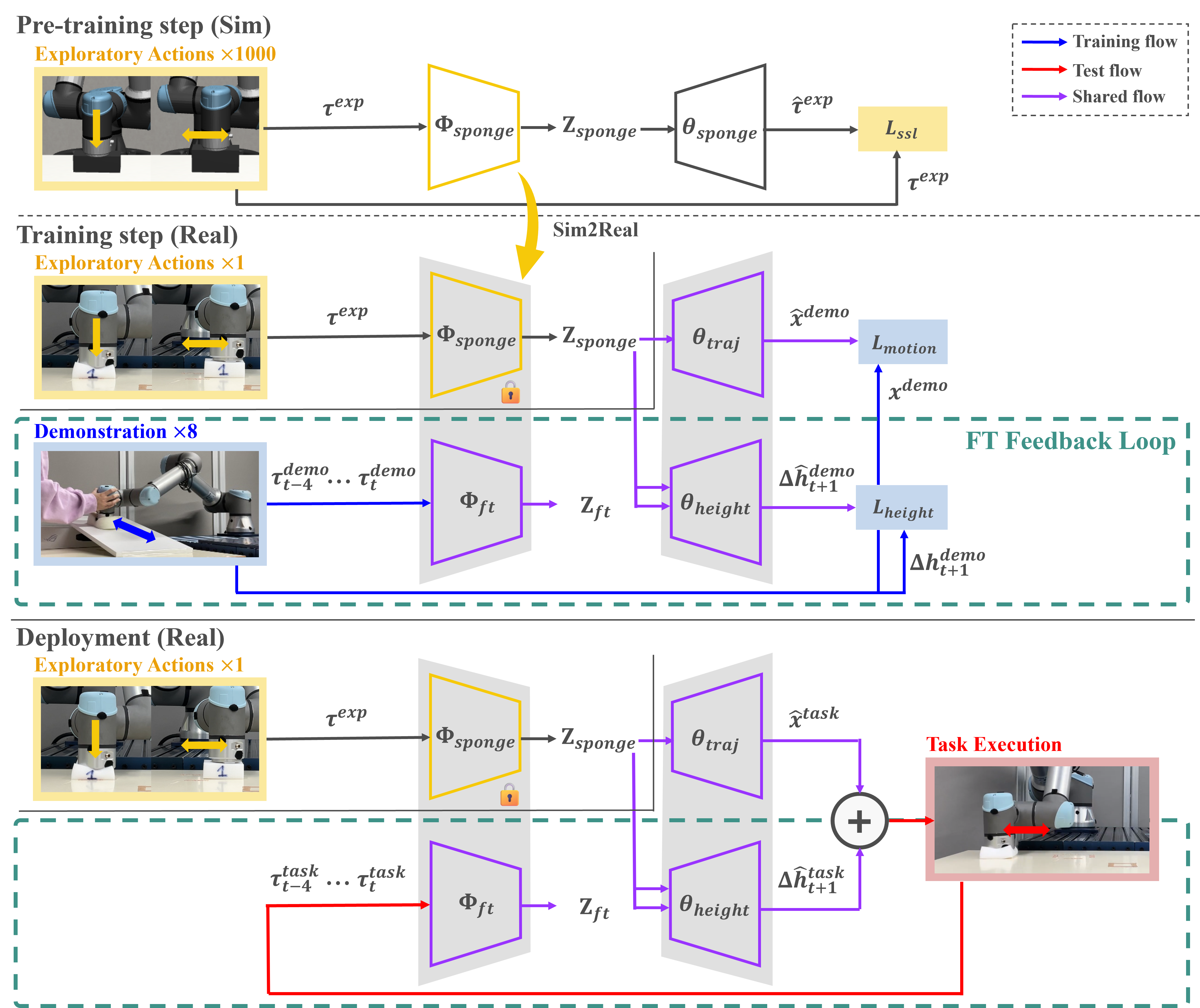}
  \caption{
  Overview of our proposed framework. First, we pre-train the sponge properties encoder $\phi_{\text{sponge}}$ using simulated unlabeled data (Pre-training step~\ref{subs:pre-training step}). Then, we train the motion trajectory decoder $\theta_{\text{traj}}$ and the FT feedback loop $\phi_{\text{ft}}-\theta_{\text{height}}$ to obtain the wiping policy with the active inference of applied force using few-shot human demonstration data (Training step~\ref{subs:training step}). Finally, we deploy the acquired policy on real robot hardware (Deployment~\ref{subs:deployment}).
  }
  \label{fig:overview}
\vspace{-7mm}
 \end{center}
\end{figure*}
\IEEEpubidadjcol
\section{Related Works and Contribution}
\label{sec:relatedworks}

Learning-based methods have succeeded in addressing contact-rich tasks. Reinforcement learning is one of the key learning-based methods for acquiring desired behaviors by defining a reward function. Martín-Martín et al. succeeded in completing a contact-rich task based on variable impedance control in end-effector space via reinforcement learning~\cite{martin2019variable}. 
Spector et al. proposed a residual admittance policy that is learned to correct the difference from the reference policy using reinforcement learning, achieving a contact-rich assembly task~\cite{spector2021learning}. However, reinforcement learning-based methods rely heavily on the reward function's design and are inefficient with few samples.

In contrast, imitation learning offers a different approach by acquiring desired behaviors through demonstrations, yielding higher sample efficiency without carefully designing the reward function. Rozo et al. used Gaussian mixture models and variable impedance control to accomplish human-robot cooperative transportation tasks~\cite{7353496}. Yamane et al. used bilateral control-based imitation learning to decouple the applied force from humans and the reaction force from the environment, enabling a robot to grasp various objects with a custom-made cross-structure hand~\cite{yamane2023soft}.

Tasks involving deformable objects are especially challenging due to the need for precise force control and adaptability to changing conditions. To address this, multiple studies combined representation learning to acquire object property embeddings prior to demonstrations. \cite{florence2019self} and \cite{DBLP:journals/corr/abs-2112-01511} pre-trained representations based on visual observations using self-supervised learning, while Guzey et al. used tactile observation for representation learning~\cite{guzey2023dexterity}. In contrast, several works suggest the importance of haptic time-series information in capturing objects’ physical properties\cite{kroemer2011learning}. 

Real-world data collection is costly and time-consuming, while simulation offers a more efficient alternative. However, Sim2Real transfer poses challenges due to differences between simulated and real-world environments. Domain randomization mitigates this gap by introducing variability in parameters like lighting and object textures during simulation, improving model robustness~\cite{weng2019DR}. Tobin et al. used domain randomization to train an object detector in simulation for robotic grasping~\cite{tobin2017domain}. Beyond visual domain randomization, dynamics randomization, which involves randomizing physical properties like mass and friction, has been explored to improve real-world generalization~\cite{peng2018sim}. Domain randomization has also been applied to manipulation tasks with deformable objects in~\cite{DBLP:conf/rss/WuYKPA20}.

Aoyama et al. used self-supervised learning on force and torque data, along with dynamics domain randomization, to capture the physical properties of deformable objects. They successfully transferred these representations from simulation to reality, enabling effective force control via few-shot imitation learning~\cite{aoyama2023fewshot}.

However, they controlled the wiping motions in an open-loop manner. Thus, the approach could not adapt to environmental changes, such as variations in wiping surface height. In contrast, methods like impedance control~\cite{hogan1985impedance} or AC~\cite{seraji1994adaptive} are well-established for closed-loop force control. However, in our context, both the target position (affected by changes in surface heights) and the target force to be applied (influenced by variations in sponge properties) are unknown, rendering these methods unsuitable. Therefore, a different approach is needed to apply the appropriate force while adapting to changes in wiping surface heights and sponge physical properties.

We addressed these challenges with three contributions:
\begin{itemize}
\item We propose a framework that combines pre-training to represent the physical properties of manipulated objects with real-time feedback of time-series force-torque (FT) information, enabling the robot’s adaptation to environmental changes from a small number of human demonstrations.
\item In contrast to the open-loop control method used in \cite{aoyama2023fewshot} for wiping motions, our approach extends this by incorporating a closed-loop control strategy. This advancement allows a robot to dynamically adapt to environmental variations, such as changes in the height of the wiping surface. Unlike admittance and impedance control methods, our approach is particularly advantageous for handling deformable and elastic objects, as it can adapt to the physical properties of unseen sponges and surface height variations without requiring prior information like target force or position.
\item We validate our approach on real hardware by altering the height of the wiping surface and the physical properties of the sponge in a wiping task, showcasing the ability to adapt to unseen environmental conditions by analyzing force measurements.
\end{itemize}

\section{Methods}

The proposed method consists of two steps: a pre-training step using a simulator and a training step using a real robot, before being deployed (Fig.~\ref{fig:overview}), each step is detailed below.

\subsection{Pre-training step} \label{subs:pre-training step}

We pre-train the sponge properties encoder $\phi_{\text{sponge}}$ on simulated unlabeled data $D_{\text{sim}} = \{(\tau^{\text{exp}})_1, \ldots, (\tau^{\text{exp}})_M\}$, collected by performing pre-defined exploratory actions (detailed in \ref{subsubs:unlabeled data}), to capture the sponges' physical properties as the latent space $Z_{\text{sponge}}$ covering a wide range of the underlying distribution. We use a self-supervised learning framework inspired by \cite{aoyama2023fewshot} but with a modified architecture. 

Using a Variational Autoencoder (VAE)~\cite{kingma2013auto} approach, the VAE encoder-decoder model $\phi_{\text{sponge}}-\theta_{\text{sponge}}$ takes FT trajectory $\tau^{\text{exp}}$ from $D_{\text{sim}}$ as inputs and outputs reconstructed FT trajectory $\hat{\tau}^{\text{exp}}$, treating the latent space $Z_{\text{sponge}}$ as a Gaussian distribution with five dimensions. 

The VAE encoder-decoder model $\phi_{\text{sponge}}-\theta_{\text{sponge}}$ consists of 2 fully connected encoder layers, 1 sampling step, and 2 fully connected decoder layers. To flatten the six sensors' time-series data $\tau^{\text{exp}} \in \mathbb{R}^{400\times6}$, we employ 2 fully connected layers each for the encoder $\phi_{\text{sponge}}$ and decoder $\theta_{\text{sponge}}$. The encoder $\phi_{\text{sponge}}$ comprises 1 fully connected layer of 5 hidden dimensions followed by the flattening step and 1 fully connected layer. Whereas the decoder $\theta_{\text{sponge}}$ comprises 1 fully connected layer with Rectified Linear Unit (ReLU) as an activation function and a dropout rate of 0.1 followed by a reshaping step and 1 fully connected layer of 5 hidden dimensions. The latent space dimension $Z_{\text{sponge}} \in \mathbb{R}^5$ is designed to capture sponges' stiffness, friction, and other non-intuitive physical properties. We adopt a loss function $L_{ssl}$  shown in Eq.~\eqref{eq:loss_ssl}, with $\beta = 0.06$.

\begin{equation}
\begin{aligned}
L_{ssl} \; &= \; E_\text{MSE}(\hat{\tau}^{\text{exp}}, \; \tau^{\text{exp}}) \hfill \\
&+ \; \beta D_\text{KL}(q_{\phi_{\text{sponge}}}(z \;|\; \tau^{\text{exp}}) \;||\; p_{\phi_{\text{sponge}}}(z)) \hfill
\end{aligned}
\label{eq:loss_ssl}
\end{equation}

\subsection{Training step} \label{subs:training step}

We train the motion trajectory decoder $\theta_{\text{traj}}$ and the FT feedback loop $\phi_{\text{ft}}-\theta_{\text{height}}$ on real-world unlabeled data
$D_{\text{real}} = \{\tau^{\text{exp}}\}$, collected by the same pre-defined exploratory actions with \ref{subs:pre-training step},
and few-shot human demonstration data
$D_{\text{demo}} = \{(x^{\text{demo}},  \Delta h^{\text{demo}},\tau^{\text{demo}})_1, \ldots, (x^{\text{demo}}, \Delta h^{\text{demo}}, \tau^{\text{demo}})_N\}$.

\subsubsection{Motion trajectory decoder $\theta_{\text{traj}}$} \label{subsubs:motion trajectory decoder}

We train the wiping motion trajectory decoder $\theta_{\text{traj}}$ using Learning from Demonstration (LfD) \cite{aoyama2023fewshot} to generate the wiping motion $\hat{x}^{\text{task}}$ according to the manipulated sponge properties. 

The encoder-decoder model $\phi_{\text{sponge}}-\theta_{traj}$ takes FT trajectory $\tau^{\text{exp}}$ from $D_{\text{real}}$ as inputs and outputs the corresponding motion trajectory $\hat{x}^{\text{demo}}$. Here, the encoder $\phi_{\text{sponge}}$ is pre-trained on simulated data $D_{\text{sim}}$, with its weights frozen during training on real data, and then deployed in the real world (Sim2Real).

The motion trajectory decoder \(\theta_{\text{traj}}\) consists of 1 fully connected layer with a dropout rate of 0.1. We adopt the Mean Squared Error $L_{traj}$ between the generated motion trajectory $\hat{x}^{\text{demo}}$ and the demonstrated one $x^{\text{demo}}$ represented in the absolute coordinate from the base link (Eq.~\eqref{eq:loss_motion}).

\begin{equation}
L_{motion} = E_\text{MSE}(\hat{x}^{\text{demo}}\,, \,x^{\text{demo}}) \label{eq:loss_motion}
\end{equation}

\subsubsection{FT feedback loop $\phi_{\text{ft}}-\theta_{\text{height}}$} \label{subsubs:tactile feedback loop}

We train an FT feedback loop $\phi_{\text{ft}}-\theta_{\text{height}}$ composed of the FT encoder \(\phi_{\text{ft}}\) and the end-effector’s vertical position decoder \(\theta_{\text{height}}\) to obtain a control input of the next time step’s vertical position according to the contact state and the manipulated sponge.

The FT encoder $\phi_{\text{ft}}$ processes the FT history from the demonstrations 
$D_{\text{demo\_ft}} = \{\tau^{\text{demo}}_{\text{t-4}},\ldots, \tau^{\text{demo}}_{\text{t}}\}$, encoding it into the latent space $Z_{\text{ft}} \in \mathbb{R}^6$, which is designed to represent the forces and torques along the x, y, and z axes. The end-effector's vertical position decoder $\theta_\text{height}$ takes the concatenated latent spaces $Z_{\text{sponge}}$ from the sponge properties encoder $\phi_\text{sponge}$ and $Z_{\text{ft}}$ from the FT encoder $\phi_{\text{ft}}$ as inputs, and outputs the next time step’s vertical displacement $\Delta \hat{h}^{\text{demo}}_{\text{t+1}}$.

The FT encoder \(\phi_{\text{ft}}\) consists of 2 layers of temporal convolutional network (TCN)~\cite{BaiTCN2018} with 25 hidden channels each and a dropout rate of 0.1. 
Inspired by~\cite{doi:10.1126/scirobotics.abc5986}, which suggests that TCN has advantages in training efficiency and training time over gated recurrent units (GRU)~\cite{chung2014empirical}, we adopt TCN as our sequence model. The end-effector’s vertical position decoder \(\theta_{\text{height}}\)  consists of 2 fully connected layers: the first fully connected layer of 128 hidden dimensions with ReLU as an activation function and a dropout rate of 0.1 followed by the final layer (the second fully connected layer).
We adopt the Mean Squared Error $L_{height}$ between the predicted vertical displacement in the next time step $\Delta \hat{h}^{\text{demo}}_{\text{t+1}}$ and that of the ground truth $\Delta h^{\text{demo}}_{\text{t+1}}$ (Eq.~\eqref{eq:loss_height}).

\begin{equation}
   L_{height} = E_\text{MSE}(\Delta \hat{h}^{\text{demo}}_{\text{t+1}}\,,\, \Delta {h}^{\text{demo}}_{\text{t+1}}) \label{eq:loss_height}
\end{equation}

\begin{figure*}[t]  
 \begin{center}

  \includegraphics[scale=0.29]{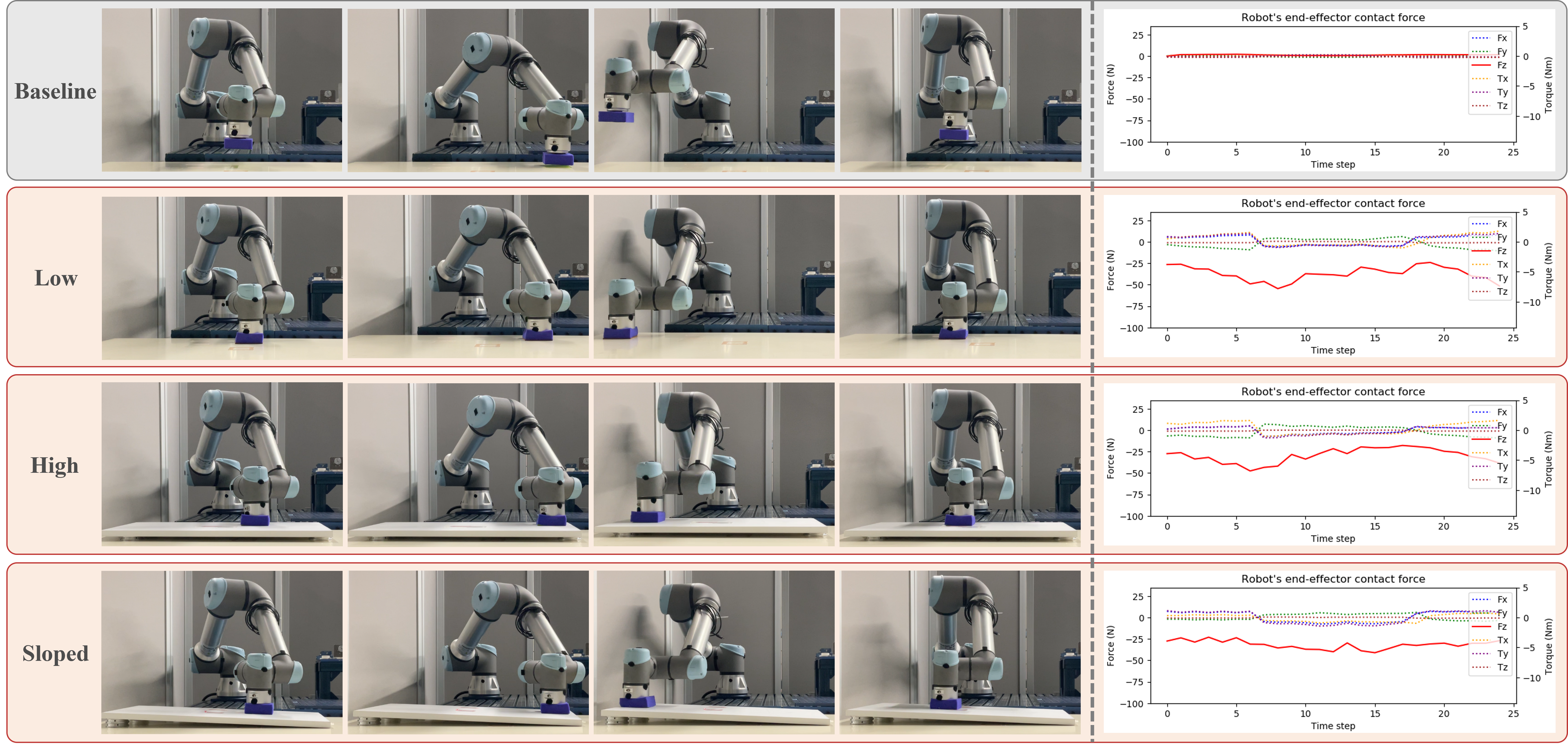}
  \caption{
   Manipulation processes of 3 different settings (low, high, sloped) using an unseen sponge that was not included in the training data. The right plots show FT profiles. The baseline simply traces the demonstration and reproduces vertical motion without considering setting changes (gray). In contrast, our method adapts to those changes while maintaining the desired wiping motion (red).
  }
  \label{fig:execution}
\vspace{-7mm}
  
 \end{center}
\end{figure*}

\subsection{Deployment} \label{subs:deployment}

In the task execution, the robot performs a wiping motion by combining offline horizontal (x, y) motion of $\hat{x}^{\text{task}}$ and online vertical (z) motion of $\Delta \hat{h}^{\text{task}}_{\text{t+1}}$. First, the robot collects unlabeled data $D_{\text{task}} = \{\tau^{\text{exp}}\}$ of the sponge being used in the task through pre-defined exploratory actions. Then it generates (x,y) planar motion $\hat{x}^{\text{task}}$ from $D_{\text{task}}$ and replays the motion offline. The FT feedback loop actively infers the next vertical position $\Delta \hat{h}^{\text{task}}_{\text{t+1}}$ from the previous $\sim$ current force and torque history $D_{\text{task\_ft}} = \{\tau^{\text{task}}_{\text{t-4}},\ldots, \tau^{\text{task}}_{\text{t}}\}$, and adapts online. 

\begin{figure}[H]
\centerline{\includegraphics[scale=0.3]{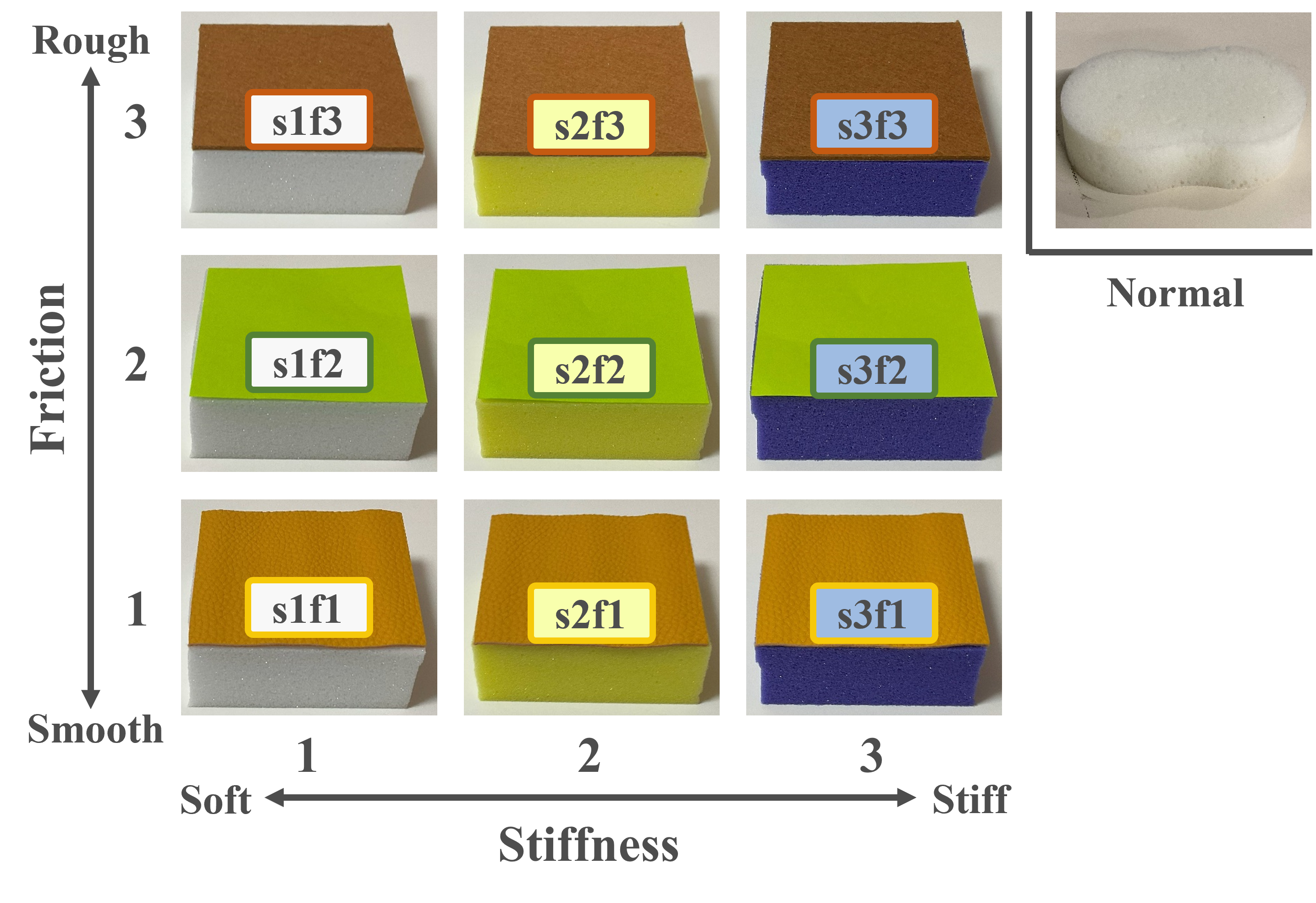}}
\vspace{-2.5mm}
\caption{10 sponges used in the experiments. One ready-made sponge (normal sponge) for training and the deployment and 9 custom-made sponges with different physical properties (3 stiffness levels $\times$ 3 friction levels)  as previously unseen sponges for the deployment. }
\label{fig:sponge}
\vspace{-1.5mm}
\end{figure}

\section{Experiment Setup}
\vspace{-1.5mm}

\subsection{Wiping task}

To illustrate our method, we use a contact-rich wiping task in which the robot has to adapt its wiping motion to the wiping surface height and the manipulated sponge's physical properties.
We prepare 3 variations of table heights (low,  high, and sloped) and 10 sponges (one ready-made sponge (normal sponge) and 9 custom-made sponges (3 stiffness levels $\times$ 3 friction levels)) as shown in Fig.~\ref{fig:sponge}. We denote a sponge with a stiffness level $m$ and a friction level $n$ as 's$m$f$n$' ($m,n = 1, 2, 3$). For additional verification, we also prepare a vertical wall to replace the horizontal table.

\begin{figure*}[t]
    \centering
    \vspace{-2.5mm}
    \captionsetup{type=table}
    \caption{Experimental results: Baselines and Ours under Various Conditions. The contact percentage indicates the proportion of time steps where force was applied to press a sponge and the number in () represents the ratio of the average force in the z-direction to that of the corresponding demonstrations (reference force) shown in Table~\ref{tab:demonstration}. }
    \label{tab:comparision}
    \resizebox{\textwidth}{!}{
    \tiny
    \begin{tabular}{|c|c|c|c|c|c|c|c|c|c|c||c|c|c|}
        \hline
        \multicolumn{2}{|c|}{\multirow{2}{*}{\diagbox{Sponge}{Height}}} & \multicolumn{3}{c|}{Low} & \multicolumn{3}{c|}{High} & \multicolumn{3}{c||}{Sloped} & 
        \multicolumn{3}{c|}{\textbf{Average}} \\
        \cline{3-14}
        \multicolumn{2}{|c|}{} & Contact & Average [N] & Std & Contact & Average [N] & Std & Contact & Average [N] & Std & 
        \textbf{Contact} & \textbf{Average [N]} & \textbf{Std}\\
        \hline
        \multirow{3}{*}{Normal} & Baseline & 12\% & 1.64 (-) & 2.38 & 32\% & -3.13 (25\%) & 8.73 & 24\% & 1.60 (-) & 4.73 & 23\% & 8\% & 5.28\\
        & AC &  100\% & -6.79 (54\%) & 1.22 & 100\% & -6.86 (54\%) & 1.02 & 100\% & -6.65 (53\%) & 4.41 & 100\% & 54\% & 2.22\\
         & Ours & \textbf{100\%} & \textbf{-13.9} (\textbf{110\%}) & 3.50 & \textbf{100\%}  & \textbf{-12.5} (\textbf{99\%})& 5.73 & \textbf{100\%}  & \textbf{-16.9} (\textbf{133\%}) & 9.12 & \textbf{100\%} & \textbf{114\%} & 6.12\\
        \hline
        \multirow{3}{*}{s1f1} & Baseline & 0\% & 1.37 (-) & 0.40 & 32\% & -1.36 (6\%) & 4.28 & 12\% & -0.14 (1\%) & 2.57 & 15\% & 2\% & 2.42\\
        & AC & 100\% & -5.84 (26\%) & 1.11 & 100\% & -5.73 (25\%) & 1.24 & 100\% & -6.68 (29\%) & 5.65 & 100\% & 27\% & 2.67\\
         & Ours & \textbf{100\%} & \textbf{-18.0} (\textbf{80\%})& 11.9 & \textbf{100\%} & \textbf{-22.1} (\textbf{97\%})& 16.7 & \textbf{100\%} & \textbf{-21.4} (\textbf{94\%})& 15.2 & \textbf{100\%} & \textbf{90\%} & 14.6\\
        \hline
        \multirow{3}{*}{s1f2} & Baseline & 0\% & 1.38 (-) & 0.36 & 12\% & -1.54 (7\%)& 6.22 & 12\% & 0.04 (-) & 2.38 & 8\% & 2\% & 2.99\\
        & AC & 100\% & -5.67 (26\%) & 0.83 & 100\% & -5.23 (24\%) & 0.77 & 100\% & -5.25 (25\%) & 6.80 & 100\% & 25\% & 2.80\\
         & Ours & \textbf{100\%} & \textbf{-28.7} (\textbf{134\%})& 18.0 & \textbf{100\%} & \textbf{-25.8} (\textbf{120\%}) & 20.1 & \textbf{100\%}  & \textbf{-23.1} (\textbf{108\%})& 17.0 & \textbf{100\%} & \textbf{121\%}& 18.4\\
        \hline
        \multirow{3}{*}{s1f3} & Baseline & 0\% & 1.28 (-) & 0.41 & 16\% & -1.12 (5\%) & 6.13 & 12\% & -0.20 (1\%) & 3.08 & 9\% & 2\% & 3.21\\
        & AC & 100\% & -5.23 (25\%) & 1.67 & 100\% & -4.94 (23\%) & 1.57 & 100\% & -6.06 (28\%) & 5.79 & 100\% & 25\% & 3.01\\
         & Ours & \textbf{100\%} & \textbf{-26.5} (\textbf{124\%})& 14.6 & \textbf{100\%} & \textbf{-20.1} (\textbf{94\%}) & 16.1 & \textbf{100\%} & \textbf{-21.5} (\textbf{101\%}) & 16.2 & \textbf{100\%} & \textbf{106\%} & 15.6\\
        \hline
        \multirow{3}{*}{s2f1} & Baseline & 0\% & 1.43 (-) & 0.38 & 40\% & -2.92 (12\%) & 7.21 & 12\% & 0.17 (-) & 2.27 & 17\% & 4\% & 3.29\\
        & AC & 100\% & -11.4 (47\%) & 1.54 & 100\% & -9.82 (41\%) & 2.17 & 100\% & -7.44 (31\%) & 7.02 & 100\% & 40\% & 3.58\\
         & Ours & \textbf{100\%}  & \textbf{-19.7} (\textbf{82\%})& 9.97 & \textbf{100\%} & \textbf{-23.3} (\textbf{97\%}) & 8.70 & \textbf{100\%} & \textbf{-15.2} (\textbf{63\%})& 4.95 & \textbf{100\%} & \textbf{81\%} & 7.87\\
        \hline
        \multirow{3}{*}{s2f2} & Baseline & 0\% & 1.32 (-) & 0.50 & 20\% & -1.58 (5\%) & 5.75 & 12\% & 0.34 (-) & 1.98 & 11\% & 2\% & 2.74\\
        & AC & 100\% & -11.7 (39\%) & 1.26 & 100\% & -10.2 (34\%) & 1.49 & 100\% & -11.3 (37\%) & 6.26 & 100\% & 37\% & 3.00\\
         & Ours & \textbf{100\%} & \textbf{-34.4} (\textbf{115\%})& 14.6 & \textbf{100\%} & \textbf{-23.3} (\textbf{77\%}) & 10.1 & \textbf{100\%} & \textbf{-21.7} (\textbf{72\%})& 7.80 & \textbf{100\%} & \textbf{88\%} & 10.8\\
        \hline
        \multirow{3}{*}{s2f3} & Baseline & 0\% & 1.76 (-) & 0.53 & 20\% & -1.46 (4\%) & 6.74 & 12\% & -0.07 (0\%) & 3.80 & 11\% & 1\% & 3.69\\
        & AC & 100\% & -12.1 (35\%) & 1.53 & 100\% & -10.7 (31\%) & 1.83 & 100\% & -11.1 (32\%) & 7.00 & 100\% & 33\% & 3.45\\
         & Ours & \textbf{100\%} & \textbf{-29.2} (\textbf{85\%}) & 14.1 & \textbf{100\%} & \textbf{-24.5} (\textbf{72\%})& 12.5 & \textbf{100\%} & \textbf{-22.2} (\textbf{65\%})& 6.86 & \textbf{100\%} & \textbf{74\%} & 11.2 \\
        \hline
        \multirow{3}{*}{s3f1} & Baseline & 0\% & 1.97 (-) & 0.62 & 20\% & -5.42 (18\%) & 12.9 & 12\% & -0.35 (1\%) & 5.83 & 11\% & 6\% & 6.45 \\
        & AC & 100\% & -18.7 (61\%) & 2.58 & 100\% & -19.5 (63\%) & 5.90 & 100\% & -16.8 (54\%) & 9.58 & 100\% & 59\% & 6.02 \\
         & Ours & \textbf{100\%} & \textbf{-30.0} (\textbf{97\%})& 15.4 & \textbf{100\%} & \textbf{-24.5} (\textbf{80\%}) & 10.5 & \textbf{100\%} & \textbf{-39.3} (\textbf{127\%}) & 13.3 & \textbf{100\%} & \textbf{101\%} & 13.1 \\
        \hline
        \multirow{3}{*}{s3f2} & Baseline & 0\% & 1.18 (-) & 0.37 & 20\% & -4.71 (13\%) & 11.4 & 12\% & -1.07 (3\%) & 5.11 & 11\% & 5\% & 5.63\\
        & AC & 100\% & -19.7 (56\%) & 3.26 & 100\% & -22.1 (63\%) & 4.24 & 100\% & -19.0 (54\%) & 8.31 & 100\% & 58\% & 5.27\\
         & Ours & \textbf{100\%} & \textbf{-37.0} (\textbf{105\%})& 8.27 & \textbf{100\%} & \textbf{-29.6} (\textbf{84\%}) & 8.31 & \textbf{100\%} & \textbf{-31.7} (\textbf{90\%})& 4.85 & \textbf{100\%} & \textbf{93\%} & 7.14\\
        \hline
        \multirow{3}{*}{s3f3} & Baseline & 0\% & 1.11 (-) & 0.46 & 44\% & -7.66 (21\%) & 17.3 & 24\% & -2.43 (7\%) & 7.24 & 23\% & 9\% & 8.33\\
        & AC & 100\% & -21.6 (59\%) & 2.35 & 100\% & -22.8 (62\%) & 2.35 & 100\% & -22.0 (60\%) & 9.61 & 100\% & 60\% & 4.77\\
         & Ours & \textbf{100\%} & \textbf{-45.2} (\textbf{123\%})& 8.97 & \textbf{100\%} & \textbf{-28.1} (\textbf{76\%}) & 9.39 & \textbf{100\%} & \textbf{-28.1} (\textbf{77\%})& 4.39 & \textbf{100\%} & \textbf{92\%} & 7.58\\
        \hhline{|==============|} 
        \multirow{3}{*}{\textbf{Average}} & \textbf{Baseline} & 1\% & 0\% & 0.64 & 26\% & 11\% & 8.67 & 14\% & 1\% & 3.90 & 14\% & 4\% & 4.40\\
        & \textbf{AC} & 100\% & 43\% & 1.74 & 100\% & 42\% & 2.26  & 100\% & 40\% & 7.04  & 100\% & 42\% & 3.68  \\
        & \textbf{Ours} & \textbf{100\%} & \textbf{106\%} & 11.9 & \textbf{100\%} & \textbf{90\%} & 11.8  & \textbf{100\%} & \textbf{93\%} & 9.97  & \textbf{100\%} & \textbf{96\%} & 11.2 \\
        \hline
    \end{tabular}
    }
    \vspace*{-4.5mm}
\end{figure*}

\subsection{Robot and Setup}

We use a 6 DoF UR5 e-series robot arm with a 6-axis FT sensor and a sponge attached to its end-effector for both simulation and real robot experiments. 
We control the robot by specifying the end-effector position when performing pre-defined exploratory actions for collecting unlabeled data and deploying our proposed method. We conduct demonstrations by moving the robot arm kinesthetically in free drive mode.
For the simulation, we use robosuite~\cite{zhu2022robosuite} with the same setup as the real robot experiments, and for controlling the real hardware, the Robot Operating System (ROS)~\cite{quigley2009ros} with the Universal Robot ROS Driver is used.

\subsection{Dataset}

\subsubsection{Unlabeled data} \label{subsubs:unlabeled data}

The robot performs two pre-defined exploratory actions \cite{aoyama2023fewshot}, where each action is carefully designed to capture the characteristics of sponges' stiffness and friction, effectively bridging the Sim2Real transfer. These actions consist of 
pressing at $0.01 m/s$ for 2 seconds and moving laterally left and right at $0.05 m/s$ for 1 second each. 
During these exploratory actions, we record the 3-axis force and torque for $4s$ at a frequency of $100 Hz$ while performing exploratory actions to obtain the FT trajectory $\tau^{\text{exp}} \in \mathbb{R}^{400\times6}$.

We collect 1000 unlabeled data in simulation for pre-training by varying sponge properties. We randomized the parameters of stiffness, friction, and damping by setting sliding friction $\mu \in [0.0, 3.5]$, solref stiffness $k \in [0.5, 1000]$ N/m, and solimp width $d \in [0.02, 0.3]$ m, to narrow the gap between simulation and reality (dynamics domain randomization).
For training, we collect 1 demonstration unlabeled data of a normal sponge. \textit{‘Normal’} refers to typical friction, stiffness, and damping properties in ready-made sponges.

The FT trajectories of the unlabeled data collected both in simulation and in the real world were similar. This is likely due to careful tuning of the dynamics-related parameters of the sponge in the simulator and the inherent elasticity of the sponge, which reduces noise in real-world measurements.

\subsubsection{Demonstration dataset} \label{subsubs:demonstration dataset}

A human demonstrator kinesthetically performs the desired wiping motion by moving the robot’s end-effector in free drive mode. 
The demonstrator is instructed to wipe the inclined table (which differs from the slope used in the validation experiments), applying as much force as possible to maximize cleaning efficiency \cite{aoyama2023fewshot}.
We collected 8 demonstrations using a normal sponge with natural speed, with no errors made by the demonstrator in completing the task.
We record the robot’s end-effector’s position, force and torque in the (x, y, z) axis at a rate of $2.5 Hz$ for $10 s$ to obtain the motion trajectory $x^{\text{demo}} \in \mathbb{R}^{25\times2}$ (2 absolute positions in (x, y) axis), vertical motion trajectory $\Delta h^{\text{demo}} \in \mathbb{R}^{25}$ (vertical displacements from the previous time step), and FT trajectory $\tau^{\text{demo}} \in \mathbb{R}^{6\times25}$.

\vspace{-2mm}
\subsection{Model training}

The datasets are pre-processed before training; we apply a Butterworth low-pass filter offline to unlabeled data and online to FT demonstrations data. 
Subsequently, we normalize all data to [0.0, 0.9].

\emph{Pre-training:}
We pre-train the sponge properties encoder \(\phi_{\text{sponge}}\) as described in \ref{subs:pre-training step} using 1000 unlabeled simulation data \ref{subsubs:unlabeled data}. We adopt the Adam optimizer and train the model for 200 epochs at a learning rate of 0.0001.

\emph{Training:}
We train the motion trajectory decoder \(\theta_{\text{traj}}\) as described in \ref{subsubs:motion trajectory decoder} using 8 motion trajectory data $x^{\text{demo}}$ of demonstrations.  We adopt the Adam optimizer as an optimizer and train the decoder for 10000 epochs at a learning rate of 0.001.
We train the FT feedback loop described in \ref{subsubs:tactile feedback loop} using 8 vertical motion trajectory data $\Delta h^{\text{demo}}$ and FT data $\tau^{\text{demo}}$ of demonstrations.  We treat FT data as time series data and set the window size as 5. We adopt the Adam optimizer as an optimizer and train the loop for 2000 epochs at a learning rate of 0.001.

\vspace{-6mm}

\section{Results and Discussion}

To evaluate our method, we conducted experiments using a robot under varying conditions in a total of 40 scenarios, including different heights of the wiping table (\ref{subs:height}) and different types of sponges (\ref{subs:sponge}). We compared our method with two state-of-the-art control methods: (1) Aoyama et al.~\cite{aoyama2023fewshot} (baseline), which is an imitation learning-based control method without an FT feedback loop, and (2) an admittance control (AC), which is a non-learning-based method. 
As noted in Section~\ref{sec:relatedworks}, in the same problem setting as the baseline and proposed method, AC cannot be executed due to the absence of necessary target force information. Therefore, during the execution of AC, we defined the target force as the force applied when the sponge is pressed by $1cm$, enabling the implementation of AC.
AC attempts to maintain this target force predicting vertical displacement $\Delta{h}$ using Eq.~\eqref{eq:ac}~\cite{Song_Yu_Zhang_2019}.  
\begin{equation}
    \Delta{h} = \frac{FT^2 + B T \, \Delta{h}_{t-1} + M \left( 2 \, \Delta{h}_{t-1} - \Delta{h}_{t-2} \right)}{M + B T + K T^2} \label{eq:ac}
\end{equation}
In this equation, $M = 0.5[Kg]$, $B = 5[N/(m/s)]$, and $K = 15[N/m]$ represent the desired inertia, damping and stiffness values, respectively. The variable $t$ denotes the $t$th sampling period, with $T = 0.4[s]$ as the sampling period. Additionally, we tested our model on a completely different setup – wiping a vertical wall instead of a horizontal table – using the same model trained with table wiping demonstration data.
For each verification, we compared the contact with the table by examining the ratio of time steps in which the sponge contacted a table. And we examined the force applied to the sponge to compare whether the robot \textit{'wiped'} with the sponge. Specifically, we used the average vertical force applied to the sponge and its standard deviation, referencing data from human demonstrations (Table~\ref{tab:demonstration}).

\begin{figure}[!t]
    \centering
    \captionsetup{type=table}
    \caption{Reference force information from Demonstrations
    }
    \label{tab:demonstration}
    \scriptsize
    \begin{tabular}{|>{\centering\arraybackslash}p{5em}|>{\centering\arraybackslash}p{5em}|>{\centering\arraybackslash}p{5em}|>{\centering\arraybackslash}p{5em}|}
        \hline
        & \multicolumn{3}{c|}{Demonstration} \\
        \cline{2-4}
         & Contact & Average [N] & Std \\
        \hline
        Normal & 100\% & -12.6 & 5.39 \\
        \hline
        s1f1 & 100\% & -22.8 & 8.11 \\
        \hline
        s1f2 & 100\% & -21.4 & 7.76 \\
        \hline
        s1f3 & 100\% & -21.3 & 10.1 \\
        \hline
        s2f1 & 100\% & -24.1 & 12.2 \\
        \hline
        s2f2 & 100\% & -30.0 & 15.8 \\
        \hline
        s2f3 & 100\% & -34.3 & 15.9 \\
        \hline
        s3f1 & 100\% & -30.9 & 9.27 \\
        \hline
        s3f2 & 100\% & -35.2 & 12.4 \\
        \hline
        s3f3 & 100\% & -36.7 & 10.5 \\
        \hline
    \end{tabular}
\vspace{-5mm}
\end{figure}

\begin{figure}[!b]
\vspace{-5mm}
\centerline{\includegraphics[scale=0.27]{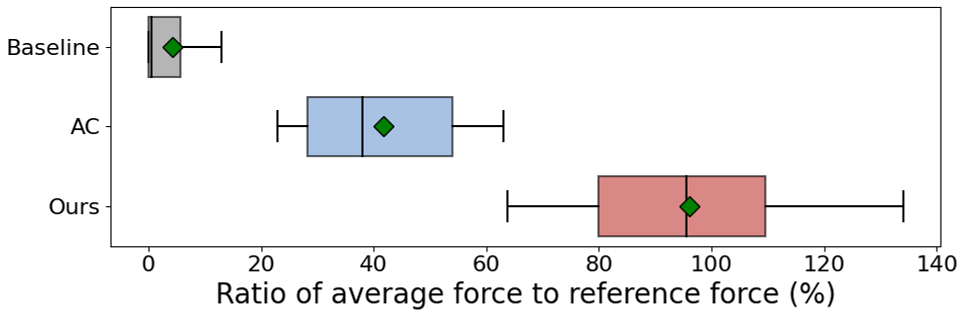}}
\caption{The boxplot illustrates the ratio of applied force to the reference force. The baseline method applied 4\% of the reference force, and the admittance control (AC) method applied 42\%. In contrast, our proposed method applied an average of 96\%, closely matching the expected value.}
\label{fig:boxplot}
\vspace{-3mm}
\end{figure}
\vspace{-2mm}

\subsection{Verification of the ability to adapt to changes in height} \label{subs:height}

We varied the wiping table’s heights (low, high, sloped) from the height used in the demonstrations (inclined table). The results are shown in Table~\ref{tab:comparision} and Fig.~\ref{fig:boxplot}. Force data during demonstration (Table~\ref{tab:demonstration}) is used as the reference force.

\begin{figure}[!t]
(a)\centerline{\includegraphics[scale=0.27]{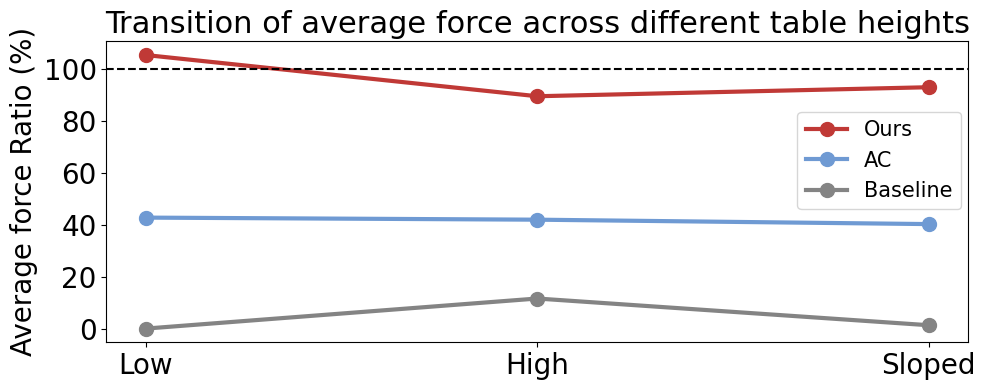}}
(b)\centerline{\includegraphics[scale=0.22]{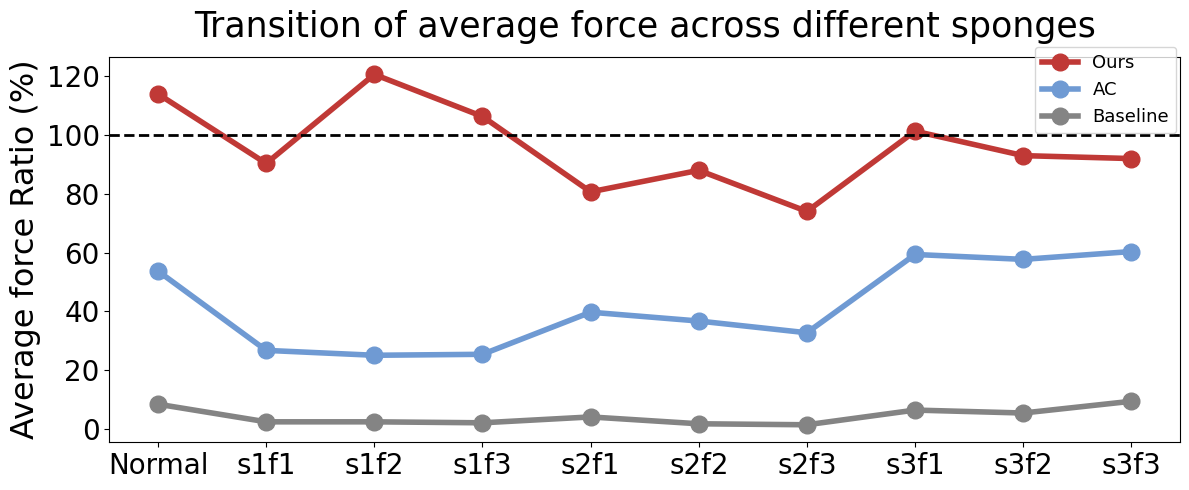}}
\caption{
    Transition in the ratio of average force compared to reference force (100\% dotted line): (a) across table heights, (b) across sponges. Our method applied force close to the expected force, outperforming the baseline and AC.}

\label{fig:heightplot}
\label{fig:spongeplot}

\vspace{-3.5mm}
\end{figure}

To adapt wiping motions to changes in the wiping surface height, the robot should apply a consistent force to the sponge regardless of the height. With the same sponge, the robot should wipe with as much force as possible to ensure effective wiping.
With the baseline method, the sponge was in contact with the table only 0-44\% of the time, and the average force reached merely 4\% of the desired reference force (Fig.~\ref{fig:boxplot}).
Specifically, in some cases with the low and sloped tables, the average force turned positive because the sponge did not contact the table, and the influence of gravitational force from the sponge's own weight became dominant. This indicates that a robot did not effectively \textit{'wipe'} and was unable to adapt to changes in the wiping surface height.
In contrast, both AC and our proposed method maintained constant contact in all 30 cases. However, AC applied only an average of 42\% of the reference force, whereas our proposed method successfully maintained an appropriate average force on the sponge across all heights, averaging 96\% of the reference force (Fig.~\ref{fig:boxplot}).
Furthermore, the applied force did not significantly vary with changes in table height as shown in Fig.~\ref{fig:heightplot} (a), with the standard deviation being only about 5\% larger than that of human demonstrations. This indicates the robot's ability to successfully adapt to height variations.

\subsection{Verification of the ability to adapt to changes in sponge} \label{subs:sponge}

We varied the sponge properties (3 stiffness levels $\times$ 3 friction levels) from the sponge used in the demonstrations (normal). The results are shown in Table~\ref{tab:comparision} and Fig.~\ref{fig:boxplot}.
Adapting the wiping motions to changes in the sponges' physical properties requires adjusting the force applied to the sponge accordingly.
With the baseline method, the robot failed to maintain contact with the table when using sponges with unseen properties. Specifically, with the low table, the contact ratio was 0\% for all 9 unseen sponges. Moreover, the average force applied was less than 25\% of the expected force, averaging only 4\% of the reference force (Fig.~\ref{fig:spongeplot} (b)). Therefore, the baseline is unable to adapt to unseen sponges.

In contrast, both AC and our proposed method successfully maintained contact at all time steps in all 30 cases. However, AC merely maintained the predefined target force without considering the sponge's physical properties, resulting in only 23-63\% of the expected force and an average of 42\% of the reference force being exerted. Our proposed method, on the other hand, applied an average force comparable to the expected force, achieving over 63\% and an average of 96\% of the reference force, according to the type of sponge (Fig.~\ref{fig:spongeplot} (b)). This demonstrates that our method successfully enables the robot to adapt to unseen sponge properties.

\begin{figure}[!b]
\vspace{-5mm}
    \centering
    \captionsetup{type=table}
    \caption{Experimental results: Wall wiping
    }
    \label{tab:wall wiping}
    \scriptsize
    \begin{tabular}{|>{\centering\arraybackslash}p{5em}|>{\centering\arraybackslash}p{5em}|>{\centering\arraybackslash}p{7.5em}|>{\centering\arraybackslash}p{5em}|}
        \hline
        & \multicolumn{3}{c|}{Wall Wiping} \\
        \cline{2-4}
         & Contact & Average [N] & Std \\
        \hline
        Normal & 100\% & -14.5 (115\%) & 2.92 \\
        \hline
        s1f1 & 100\% & -23.7 (104\%) & 17.1 \\
        \hline
        s1f2 & 100\% & -29.5 (138\%) & 21.6 \\
        \hline
        s1f3 & 100\% & -25.4 (119\%) & 19.5 \\
        \hline
        s2f1 & 100\% & -29.0 (120\%) & 19.4 \\
        \hline
        s2f2 & 100\% & -32.2 (107\%) & 19.7 \\
        \hline
        s2f3 & 100\% & -28.9 (84\%) & 18.4 \\
        \hline
        s3f1 & 100\% & -27.0 (87\%) & 13.8 \\
        \hline
        s3f2 & 100\% & -33.5 (95\%) & 17.1 \\
        \hline
        s3f3 & 100\% & -27.3 (74\%) & 19.2 \\
        \hline
    \end{tabular}
\vspace{-6mm}
\end{figure}

\vspace{-4.5mm}

\subsection{Wall Wiping}
\begin{figure}[!t]
\vspace{-1.5mm}
\centerline{\includegraphics[scale=0.22]{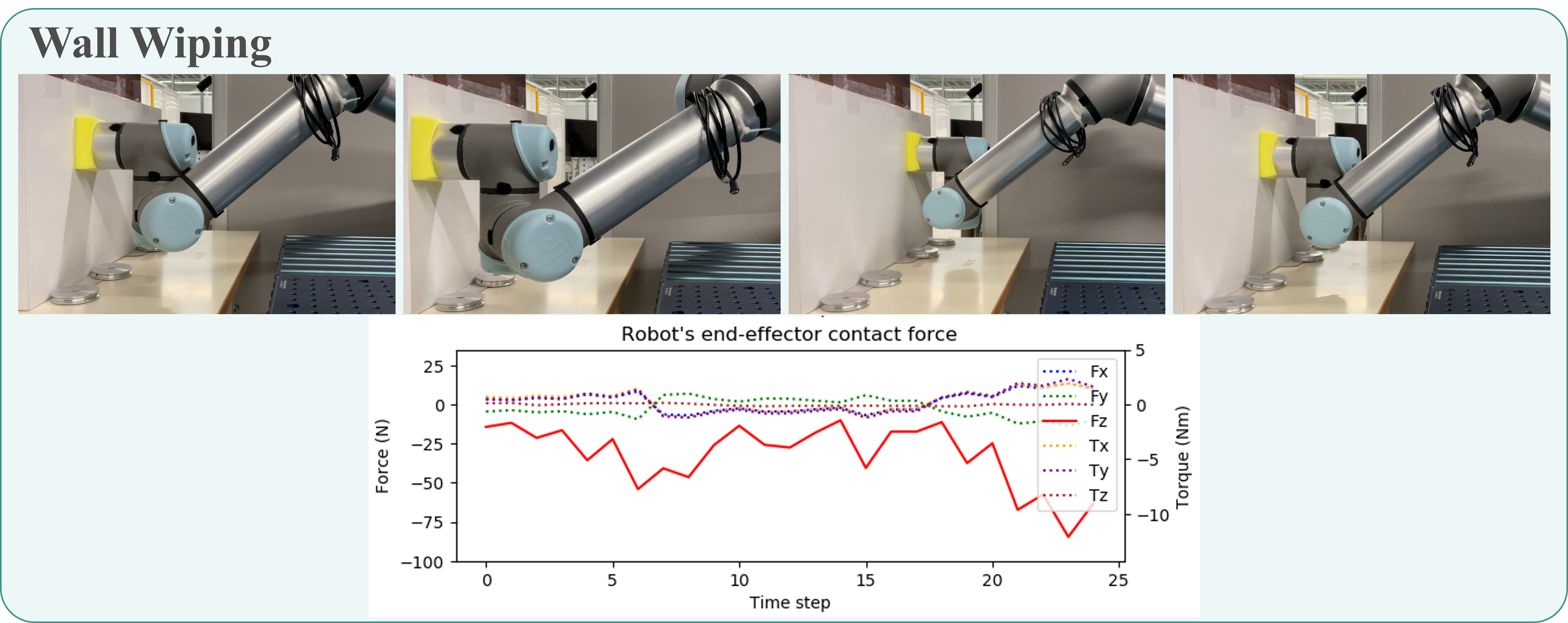}}
\caption{Wall wiping task using an unseen sponge. The plots show FT profiles. Our method enables a robot to wipe a vertical wall with table wiping data.}
\label{fig:task}
\vspace{-3mm}
\end{figure}

In real-world scenarios, cleaning involves more than just wiping horizontal surfaces like tables; it may include tasks such as wiping walls and other vertical surfaces. A key challenge for robots in these tasks is the ability to adapt to the physical properties of sponges and adjust the applied force in real time as surface conditions change. Our method achieves this adaptiveness independently of gravitational effects. In previous tasks (\ref{subs:height} and \ref{subs:sponge}), the direction of the forces applied to the sponge was aligned with gravitational acceleration, whether the configuration was low, high, or sloped. To further demonstrate that our method is effective regardless of gravity's influence, we tested our method in a gravity-neutral setting—wall wiping—where gravitational forces do not affect the applied forces during the task.

We evaluated the same model as \ref{subs:height} and \ref{subs:sponge}, trained using the same demonstration data of table wiping. Due to the setting changes, the end-effector’s frame rotated 90 degrees and the base-link’s x-axis came vertically to the end-effector. We swapped the position outputs of the x-axis and z-axis based on the base-link, and introduced an offset to the z-axis positions.  Although this might appear as a mere transformation of output trajectories, the core challenge lies in the method’s ability to adjust applied forces in a gravity-independent manner. The results are shown in Table~\ref{tab:wall wiping}.  

Our method maintained contact with a wall in all 10 cases and the applied forces were comparable to that expected, averaging 104\% of the reference force. This indicates that a robot can adapt to wall wiping even with unseen sponges.
\vspace{-2.5mm}

\section{Conclusion}

This work tackles the challenges of robots adapting to environmental changes in manipulating deformable objects in contact-rich tasks with few human demonstrations.  Our method combines real-time FT feedback with pre-trained object representations in closed-loop by treating contact information as time series data. Focusing on a wiping task, we varied table heights and sponge properties. To verify the effectiveness of the proposed method, we also tested the proposed method on a wall-wiping task. Experimental results show that the robot adapts to unseen manipulating surface height and object properties with our method, surpassing performances of the baseline and AC methods. 

Although we demonstrated our approach’s adaptiveness, we found that the standard deviations were $5 \%$ greater on average than that of the human demonstrations, and about 3 times greater compared to AC. 
This increased variability suggests that further refinement of the control policy is needed to achieve more consistent results, which is crucial for tasks requiring high precision and consistency but probably enough for daily life tasks.
Our method has another limitation: it is designed specifically for tools that are deformable and elastic. 
The approach follows this premise in \cite{aoyama2023fewshot}, where applying as much force as possible maximizes cleaning efficiency.
However, when the robot attempts to wipe with rigid objects (e.g., bricks), our method applies excessive force according to the object's hardness, causing the robot to trigger safety alarms and stop. In contrast, admittance and impedance control methods can handle such cases by adjusting target force and position.

While our method is more advantageous than admittance and impedance control for deformable and elastic objects, future work will focus on expanding its applicability to a wider range of objects, including non-deformable ones. One idea is to enhance our system by pre-training it on a large dataset, similar to models like PaLM-E \cite{driess2023palme}, so the robot can adjust its actions based on visual or linguistic prompts. This would allow to create personalized motion for different object types. 
Our method would be more versatile and adaptive across various real-world scenarios.

\begin{table}[!t]
\centering
\caption{The average ratio of the applied force in the z-direction to the reference force exerted by the demonstrator.}
\resizebox{\columnwidth}{!}{%
\scriptsize 

\begin{tabular}{l|cc|cc|cc||c}
\hline
                & \multicolumn{2}{c|}{Layer} & \multicolumn{2}{c|}{Window Size} & \multicolumn{2}{c||}{Demo} & \multirow{2}{*}{Standard} \\ \cline{2-7}
                & Fewer  & More  & Smaller  & Larger  & Fewer  & More  &  \\ \hline
Average (\%)    & 159    & 152   & 182      & 170     & 190    & 114   & \textbf{97} \\ \hline
\end{tabular}%
}

\label{tab:ablation}
\vspace{-3mm}
\end{table}
\appendix

We conducted ablation studies to validate: (1) the number of layers in the FT feedback loop, (2) the window size of TCN, and (3) the number of demonstrations. Additionally, pre-training ablation studies have been conducted in ~\cite{aoyama2023fewshot}, demonstrating its impact on improving the generation of desired wiping motion for unseen sponges.
In each ablation study, we used the values from our proposed method as the standard and compared them with two variants: smaller values and larger values. We evaluated the model’s performance by comparing the force exerted in z-direction by the robot with the reference force exerted by a human during the demonstration. We tested the system under 2 wiping surface heights (low and high) and 2 sponge types—a known sponge (Normal) used in the demonstration and an unknown sponge (s2f1) not used during the demonstration—resulting in $2 \times 2 = 4$ combinations. The results are shown in Table~\ref{tab:ablation}.

\begin{figure}[!b]
\vspace{-5mm}
(a)\centerline{\includegraphics[scale=0.32]{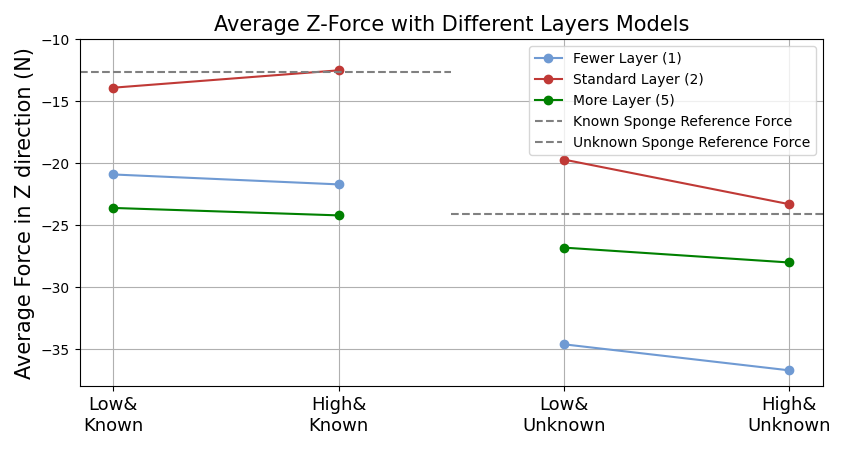}}
(b)\centerline{\includegraphics[scale=0.32]{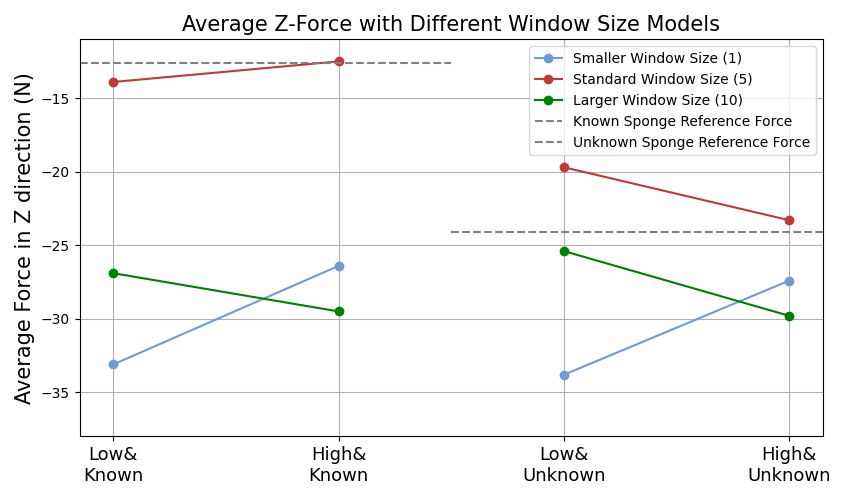}}
(c)\centerline{\includegraphics[scale=0.32]{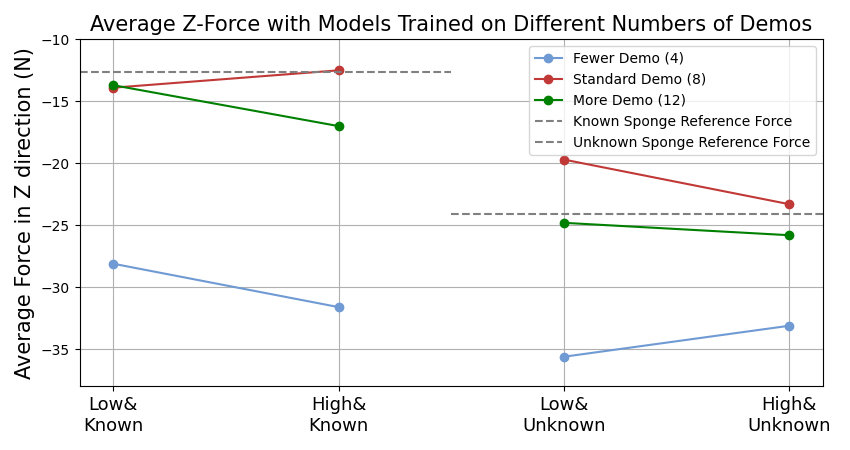}}
\vspace{-2.5mm}
\caption{
Ablation study results: (a) Number of layers in the FT feedback loop.  (b) Window size of TCN. (c) Number of demonstrations. The two dashed lines represent the reference force exerted by the demonstrator for each sponge.} 

\label{fig:layer}
\label{fig:window}

\label{fig:demo}
\vspace{-4.5mm}
\end{figure}

\emph{Number of layers in the FT feedback loop:}
We compared the proposed 2-layer model with a fewer-layer model (1 layer) and a more-layer model (5 layers) to analyze the effects of model depth on performance (Fig.~\ref{fig:layer} (a)). In the fewer-layer model, although the force applied to the sponge was adjusted, the reference force was 1.43 to 1.73 times higher, showing that the model lacked sufficient capacity to capture the necessary force control dynamics. On the other hand, the more-layer model appeared to handle the unknown sponge well at first glance but applied a force around -25N regardless of the sponge’s physical properties. This suggests that the deeper model was too complex and failed to learn force control dynamics and generalize well.  
These results indicate that neither too shallow nor too deep a model performs well, with the optimal performance achieved at a depth of around 2 layers, where the balance between model capacity and complexity is effectively maintained.

\emph{Window size of TCN:}
We compared the proposed window size of 5 with a smaller window size (window size of 1) and a larger window size (window size of 10) to analyze the effects of the length of the past history referenced on performance (Fig.~\ref{fig:layer} (b)). The smaller window size model applied approximately $-33N$ for low surfaces and $-27N$ for high surfaces, regardless of the sponge type. The difference in force applied with the change in surface height exceeded 5N, indicating that the model could not handle either the wiping surface height or sponge properties. 
These suggest that the window size should neither be too short nor too long. For this wiping task, a window size of 5 is appropriate.

\emph{Number of demonstrations:}
We compared the proposed model with 8 demonstrations against a fewer-demo model (4 demonstrations) and a more-demo model (12 demonstrations) to analyze the effect of demonstration quantity on performance (Fig.~\ref{fig:layer} (c)). The fewer-demo model exerted excessive force of $-30N$ in all conditions, suggesting that it learned to simply push hard regardless of the conditions. The more-demo model showed only small changes in applied force when the surface height changed and applied forces close to the reference force when the sponge type changed. The performance was nearly identical to the standard model. As shown in Table~\ref{tab:ablation}, the average ratio of applied force to reference force for the more-demo model (114\%) was similar to the standard model (97\%). Therefore, a minimum of 8 demonstrations is sufficient for the model to learn the relationship between the FT history and the end-effector’s next position.

\bibliographystyle{IEEEtran}
\bibliography{reference}

\end{document}